\def\year{2016}\relax
\documentclass[letterpaper]{article}
\usepackage{aaai16}
\usepackage{times}
\usepackage{helvet}
\usepackage{courier}
\usepackage{amsmath}
\usepackage{amssymb}

\usepackage{times}
\usepackage{latexsym}
\usepackage{graphicx}
\usepackage{amssymb,amsmath,bm}
\usepackage{textcomp}
\usepackage{array}
\usepackage[T2A,T1]{fontenc}
\usepackage[utf8]{inputenc}
\usepackage[russian,english]{babel}
\usepackage{algorithm}
\usepackage{algpseudocode}
\usepackage{graphicx,multicol}
\usepackage{subcaption}
\usepackage{multirow}
\title{Feasibility of Post-Editing Speech Transcriptions with a Mismatched Crowd}
\author{Purushotam Radadia, Shirish Karande\\
Tata Research Development and Design Center, TCS-Innovation Labs, Pune, India\\
  purushotam.radadia@tcs.com, shirish.karande@tcs.com}

\begin{document}

\maketitle
\begin{abstract}
Manual correction of speech transcription can involve a selection from plausible transcriptions. Recent work has shown the feasibility of employing a mismatched crowd for speech transcription. However, it is yet to be established whether a mismatched worker has sufficiently fine-granular speech perception to choose among the phonetically proximate options that are likely to be generated from the trellis of an ASRU. Hence, we consider five languages, Arabic, German, Hindi, Russian and Spanish. For each we generate synthetic, phonetically proximate, options which emulate post-editing scenarios of varying difficulty. We consistently observe non-trivial crowd ability to choose among fine-granular options.
\end{abstract}
\section{Introduction}
The ASR errors are unavoidable, especially in case of low-resourced languages. Human intervention is required to correct such errors. Crowds can be utilized for post-editing ASR transcripts \cite{mcgraw2010collecting,Wald:2011}. The Scribe system proposed in \cite{lasecki2012real} allows non-experts to caption the speech in near real time (less than 5 sec latency). The study reported in \cite{gaur2015effects} analyzes latency of transcriptions achieved from crowd based on the errors in an ASR output.  
In above mentioned studies, a worker must be familiar with the spoken language. This can limit the crowd base for under-resourced languages due to the significant mismatch between population of native speakers of languages in the world and the crowd workers who speak that language \cite{pavlick2014language}. Therefore, in \cite{jyothi2015acquiring,purushotam2016} the use of a mismatched crowd, which is unfamiliar with the source language has been explored for transcription. 
Nevertheless, to the best of our knowledge, the utility of mismatched crowd is not explored for transcription post editing. Our contribution in this paper is to employ a mismatched crowd that performs transcription voting tasks for a given utterance.  We consider five different languages in our study: Arabic, German, Hindi, Russian, Spanish languages. 

\section{Corpus Creation}
{\bf Word Selection:} We select  $500$ words each from, viz., Arabic, German, Hindi, Russian and Spanish. The words are sampled from publically available pronunciation dictionaries that are being used in ASR systems\footnote{https://sourceforge.net/projects/cmusphinx/}. 
The pronunciation dictionary contains words and their phonetic decompositions in terms of arpabets.  
The words are selected in such a way that the overall corpus becomes phonetically rich for each language. In addition, the length of selected words is limited to 4-8 arpabets. 
%
      %
 
{\bf Synthetic Utterances:} Since, the word utterances in natural speech can be dependent on the speaker as well as the acoustic conditions of the recording environment, we use synthetic utterances in our experiments. We synthesize language specific word utterances using Google speech synthesis interface via SoundOfText\footnote{http://soundoftext.com/} utility. 

{\bf Generation of Synthetic Options:}
%
Since, we are interested in studying the ability of the crowd to capture the fine grained phonetic information, the valid dictionary word may not provide the proper set of options which are phonetically proximate. Hence, we aim to generate them synthetically.  
We introduced phonetic substitution errors to generate the perturbed sequence. Further, we only allow the vowel to vowel and consonant to consonant substitutions and not cross substitutions. This helps in maintaining the pronunciability of the word.
We present a worker with a set of 5 options (4 strings and a "none of the above"). The 4 sequences are generated from the reference sequence to form an option set for a given word.  We considered 6 such option sets for each word in the corpus. The sets are, $S_{0124}$, $S_{0112}$, $S_{0111}$, $S_{1234}$, $S_{1124}$ and $S_{1111}$.
This results in 24 different sequences for each reference sequence. The set $S_{0124}$ indicates that the alternate arpabet sequences are 0, 1, 2 and 4 substitutions away from reference, respectively. 

{\bf Phoneme to Grapheme Modeling:}
Since the crowd worker may neither be familiar with the phone set nor the script of a spoken language, 
we convert all the phoneme sequences to the valid script that workers are familiar with. We chose Roman(English) script for this purpose. To transform the phoneme sequence to the Roman letters (Graphemes), Phoneme to Grapheme models (P2G) models are used. 
We used Carmel Finite State Toolkit \cite{carmel} to build P2G model for each language. 
Consequently, the trained P2G models enable us to create the Roman letter sequences of the perturbed arpabet sequences of the alternate list.

{\bf Task Allocation and Crowd Work:}
We used the CrowdFlower platform for our experiments. We created 6 tasks for each of the 500 words from the considered 5 languages. Each task was repeated thrice to study vote aggregation, as well as remove spam bias. Each worker was allocated 15 randomly selected tasks and thus we employed 3000 workers to complete 15000 voting tasks.

\section{Experimental Results}
{\bf Crowd Bias against Rejection:}
We observed that crowd distribution is skewed towards the quadrant where accuracy if true world listed is greater than 50\%, while the accuracy if word is not-listed is less than 50\%. This indicates that majority of the crowd fails to exercise the Reject option when similar sounding options are available in the list. Hence, in the remainder of the paper, we focus on the cases where the correct word is always present in the list shown to the worker.

	
	{\bf Performance of Crowd workers:}
	Table~\ref{tab:workeracc} shows that as the task difficulty increases (i.e., the options becomes more phonetically similar), the performance is degraded. We highlight that even an accuracy of 45\%, as observed for Arabic over the hardest options, is significantly better than that of random guess accuracy of 20\% (or 25\%). Meanwhile, for the Indo-European languages the average accuracy, especially for the relatively easy tasks ($S_{0124}$) is greater than 75\%. 
	
%
\begin{table}[b]
\vspace{-0.5cm}
        \centerline{
          \begin{tabular}{| m{1.2cm} | m{1.2cm} | m{1.2cm} | m{1.2cm} |}
            \hline
           \bf & \bf $S_{0124}$ & \bf $S_{0112}$  & $S_{0111}$\\
					 \hline
           ar & $0.54$& $0.48$ & $0.45$\\
            \hline
						de & $0.66$&$0.6$ & $0.58$\\
            \hline
						hi & $0.7$& $0.63$ & $0.61$\\
            \hline
						ru & $0.68$& $0.61$ & $0.59$\\
            \hline
						es & $0.71$& $0.65$ & $0.62$\\
            \hline
          \end{tabular}
        }
				\vspace{-0.2cm}
			\caption{\label{tab:workeracc} {\it Voting accuracies across languages}}
      \end{table}
			

		{\bf Language Tree Analysis:}
	The native language of a transcriber can significantly impact his/her perception of the mismatched speech. Table \ref{tab:langacc} shows the performance of workers, belonging to certain languages/language-families.  The language families considered in the Table \ref{tab:langacc} are represented as: Indo-Aryan=\{Bengali, Bihari, Urdu, Sinhalese, Gujarati, Marathi, Nepali\}, Slavic=\{Polish, Czech, Croatian, Ukranian, Bulgerian, Serbian, Slovenian, Slovak \}, Romance=\{Italian, Portugese, French\} and Tonal=\{Chinese, Vietnamese\}. 
It is evident that for each of the five languages the best accuracies were exhibited by the natives. The fact that natives are unable to provide near 100\% accuracy can be attributed to spam, but, also to the fact that task involves utterances from multiple languages, thus, denying the listener of a context. In addition, one can observe some correlation across related languages, for example, the second best performance for Hindi was by the Indo-Aryan group, for Russian was by the Slavic group and for Spanish was by the  Romance group. We could not establish such a pattern for German (where English should have been expected to be the second best.) The correlations in speech perceptions may indeed be influenced by modern cultural influences and shared words, rather than just linguistic connections. 
	
	\begin{table}[t]
        \centerline{
          \begin{tabular}{| m{2cm} | m{0.6cm} | m{0.6cm} | m{0.6cm} | m{0.6cm} | m{0.6cm} |}
            \hline
           \bf & \bf ar & \bf de  & \bf hi & \bf ru  & \bf es \\
					 \hline
           Arabic & $\bf 0.59$& $0.56$ & $0.62$ & $0.59$ & $0.61$\\
            \hline
						Turkish & $0.5$&$0.58$ & $0.65$ & $0.57$ & $0.57$\\
            \hline
						German & $0.42$& $\bf 0.77$ & $0.67$ & $0.57$ & $0.57$\\
            \hline
						English & $0.47$& $0.61$ & $0.67$ & $0.61$ & $0.63$\\
            \hline
						Hindi & $0.51$& $0.51$ & $\bf 0.75$& $0.64$ & $0.62$\\
            \hline
						Indo-Aryan & $0.48$& $0.54$ & $\bf 0.74$& $0.56$ & $0.58$\\
            \hline
						Russian & $0.56$& $0.61$ & $0.59$& $\bf 0.78$ & $0.61$\\
            \hline
						Slavic & $0.48$& $0.64$ & $0.63$& $\bf 0.65$ & $0.64$\\
            \hline
						Spanish & $0.48$& $0.59$ & $0.63$& $0.62$ & $\bf 0.76$\\
            \hline
						Romance & $0.51$& $0.64$ & $0.66$& $0.62$ & $\bf 0.68$\\
            \hline
						Tonal & $0.40$& $0.49$ & $0.62$& $0.65$ & $0.46$\\
            \hline
					\end{tabular}
        }
				\vspace{-0.2cm}
			\caption{\label{tab:langacc} {\it Impact of worker's language on Accuracies}}
			\vspace{-0.5cm}
      \end{table}
	
		{\bf Feasibility of Aggregating Labels:}
We consider two very simple vote aggregation strategies. The first one is the traditional majority vote decoding, where, the aggregated label is the majority label if a simple majority exists, else if all three labels are distinct, we simply declare an error. The second algorithm, seeks to break such three way ties, by building worker reputations on the tasks where a majority does exist. The reputation is described by the count of correct votes in simple pure majorities. In a three-way tie, the worker with the highest such repute is declared a winner. 
Table \ref{tab:majvotb} shows the performance of the Marjority Vote (MV) as well as the Majority Vote with Tie-Breaks (TB). When compared to Table \ref{tab:workeracc}, MV provides a gain in accuracy of (1-2)\% for Arabic, (2-4)\% for German, (3-6)\% for Hindi, (3-4)\% for Russian and (2-5)\% for Spanish. Additionally, it can be observed that reputation based TB provides an additional gain over MV of (4-5)\% for Arabic, (4-5)\% for German, (2-5)\% for Hindi, (3)\% for Russian and (3-4)\% for Spanish. 

			
			\begin{table}[b]
			\vspace{-0.5cm}
			\centerline{
  \begin{tabular}{|l|l|l|l|l|l|l|}
    \hline
      &\multicolumn{2}{c|}{\bf $S_{0124}$} &
      \multicolumn{2}{c|}{\bf $S_{0112}$} &
      \multicolumn{2}{c|}{$S_{0111}$} \\
			\hline
    & MV & TB & MV &TB & MV & TB \\
    \hline
    ar & $0.55$ & $0.60$ & $0.49$ &  $0.53$& $0.47$ &$0.52$\\
    \hline
    de & $0.7$ & $0.74$ & $0.63$ &  $0.68$&  $0.60$&$0.65$\\
    \hline
    hi & $0.76$ & $0.78$ & $0.68$ &  $0.72$& $0.64$ &$0.69$\\
    \hline
		 ru & $0.74$ & $0.77$ & $0.64$ & $0.67$ & $0.63$ &$0.66$\\
    \hline
		 es & $0.73$ &$0.77$ & $0.70$ & $0.73$ &  $0.67$&  $0.70$\\
    \hline
  \end{tabular}
	}
	\vspace{-0.2cm}
	\caption{\label{tab:majvotb} {\it Majority vote with tie breaking}}
\end{table}
	%
			
\section{Conclusion}
It was observed that a mismatched crowd is reluctant to use a Reject Option, in future, we may have to provide improved interaction to un-bias the crowd.  However, in the cases where the true value is a part of the option list it was observed that a mismatched crowd can indeed identify the correct option even among fine-granular (phonetically proximate) choices. There is early evidence that some language pairs (or demographies) though non-native may be better matches, similarly there is evidence that the voting errors are not correlated as shown by the gain in accuracy through even majority voting based strategies. 

\bibliographystyle{aaai}
\bibliography{hcomp16_cam2}

\end{document}